\documentclass[journal]{IEEEtran}

\usepackage{cite}
\usepackage{amsmath,amssymb,amsfonts}
\usepackage{algorithmic}
\usepackage{graphicx}
\usepackage{textcomp}
\usepackage{subfigure}
\usepackage{xcolor}
\usepackage{array}

\DeclareGraphicsExtensions{.png,.pdf}

\begin{document}

\title{SPECT Imaging Reconstruction Method Based on Deep Convolutional Neural Network}

\author{Charalambos Chrysostomou$^{*}$,
        Loizos Koutsantonis,~\IEEEmembership{Member,~IEEE,}
        Christos Lemesios,~\IEEEmembership{Member,~IEEE,}
        and Costas N. Papanicolas,
\thanks{C. Chrysostomou, L. Koutsantonis, C. Lemesios and C.N. Papanicolas are with the computation-based Science and Technology Research Center, The Cyprus Institute, 20 Konstantinou Kavafi Street, 2121, Aglantzia, Nicosia, Cyprus}
\thanks{*Corresponding Author}%

}

\IEEEpubid{978-1-7281-4164-0 /19/\$31.00 ~\copyright2019 IEEE}

\maketitle



%
\IEEEpeerreviewmaketitle

\section{Abstract}

In this paper, we explore a novel method for tomographic image reconstruction in the field of SPECT imaging. Deep Learning methodologies and more specifically deep convolutional neural networks (CNN) are employed in the new reconstruction method, which is referred to as "CNN Reconstruction – CNNR". For training of the CNNR Projection data from software phantoms were used. For evaluation of the efficacy of the CNNR method, both software and hardware phantoms were used. The resulting tomographic images are compared to those produced by filtered back projection (FBP) \cite{bruyant2002analytic}, the “Maximum Likelihood Expectation Maximization” (MLEM) \cite{bruyant2002analytic} and ordered subset expectation maximization (OSEM) \cite{hudson1994accelerated}.

\IEEEpubidadjcol

\section{Introduction}

Single Photon Emission Computerized Tomography (SPECT) \cite{wernick2004emission, madsen2007recent, mariani2010review} and Positron Emission Tomography (PET) \cite{bailey2005positron, vaquero2015positron}, have a key role in emission tomography and medical imaging as being the key methods. All emission tomography techniques function by detecting the concentrations of the isotope tagged to a biochemical compound injected into the body. These compounds are absorbed by organs at varying rates according to their affinity to the biochemical compound \cite{reader20144d}. In the standard SPECT system, up to three detector heads are used \cite{vanzi2004kinetic}, that revolves around the body and can detect isotope decay in the organs of interest. In all emission tomography methods, the rate of absorption, scattering effects and the background radiation \cite{niu2011effects, ritt2011absolute, frey1994modeling} can affect the quality of the reconstruction. High doses of radiopharmaceuticals are needed to improve image reconstruction's quality, which can have negative impacts on the health of the patients while reducing these doses, limits the image reconstructions and statistics. Thus a new method is needed that can perform high-quality image reconstructions while maintaining the radiopharmaceuticals doses to a minimum. In this paper, based on previous work \cite{chrysostomou2018reconstruction}, a novel method is developed and presented that utilises convolutional neural networks to perform tomographic image reconstructions.

The paper is organized as follows: Section \ref{sec:training_data} presents the generated training data for the proposed model, Section \ref{sec:cnn}, introduces the proposed model, Section \ref{sec:results}, presents the results and finally Section \ref{sec:conclussion} is discussion and conclusions.

\section{Training Data}
\label{sec:training_data}

\IEEEpubidadjcol

\begin{figure*}[htp]
    \centering
        \includegraphics[width=15cm]{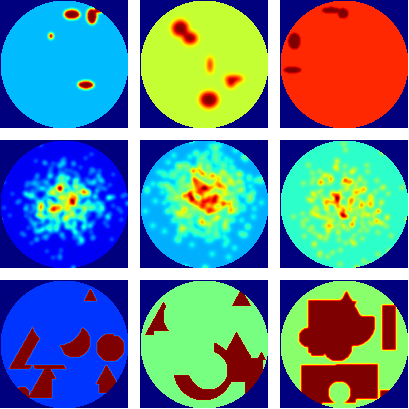}
\caption{Random samples of the generated images used to train the CNNR proposed method}
\label{fig:random_phandoms}
\end{figure*}

In order to train the proposed method, 600,000 software phantoms were used, generated randomly. For each randomly generated phantom of $128 \times 128$ pixels, sinograms (vectorised projections) were obtained through 
\begin{equation}
  Y_{i} = \sum_{j=1}^{N\,PxN\,R} P_{ij} F_{j}
  \label{eq:1}
\end{equation}

\noindent where $N\,P$ is the number of projection angles, in this case, 128 projections, equally spaced in 360 degrees, and $N\,R$ is the number of bin measurements per projection angle. The acquired sinograms were additionally randomised with a Poisson probability distribution to provide the noisy sets of projections. Three levels of Poisson noise was used by scaling the values of the sinograms to 90\%, 50\%, and 10\%, as Low, Medium and High noise. Examples of the random phantom generated can be found in figure \ref{fig:random_phandoms}

\begin{figure}[htp]
    \centering
        \includegraphics[width=6cm]{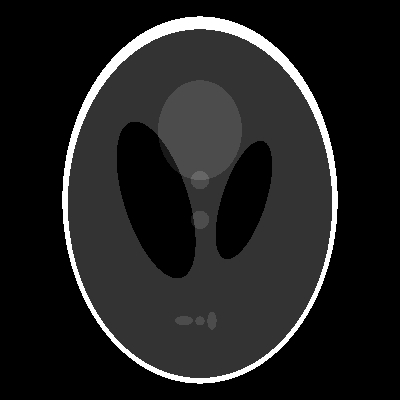}
\caption{Shepp Logan Phantom used to evaluate and demonstrate the capabilities of the proposed method}
\label{fig:shepp_logan}
\end{figure}

Finally, to assess and evaluate the capabilities of the proposed method, the Shepp Logan phantom \cite{shepp1974fourier} was used as showed in figure \ref{fig:shepp_logan}.

\begin{figure*}
    \centering
    \includegraphics[width=14cm]{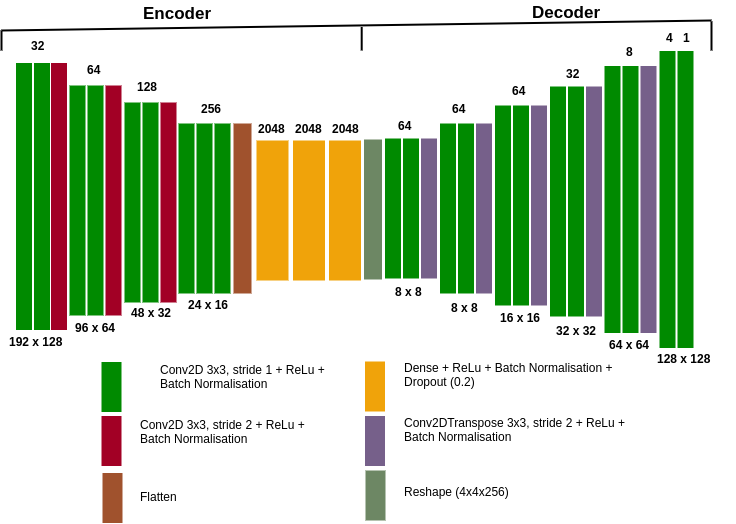}
    \caption{Proposed Deep Convolutional Neural Network Model for the SPECT image reconstruction}
    \label{fig:proposed_cnn_model}
\end{figure*}

\subsection{Convolutional Neural Network}
\label{sec:cnn}

The proposed method employed convolutional neural network (CNN) which is a is a deep feed-forward artificial neural networks subtype. CNN's have been employed in recent years to parse visual representations with many applications such as image classification and recognition \cite{li2014medical}, recommender systems \cite{cheng2016wide} and medical image analysis \cite{milletari2016v}. The advantage of employing CNN's is that they need relatively less preprocessing and manipulation of data and prior knowledge, in comparison to other existing methodologies. Thus the advantage of CNN's is that they can be employed with no prior knowledge and input from experts.  The proposed model consists of two parts, the encoder, and decoder, as shown in Figure \ref{fig:proposed_cnn_model}. The input of the model, the sinograms, is of size 192 x 128 x 1 and the output of the model as the original "true" activity the size of 128 x 128 x 1.

The proposed model consists of two parts, the encoder and decoder, as shown in Figure \ref{fig:proposed_cnn_model}. The input of the model, the sinograms, is of size 192 x 128 x 1 and the output of the model as the original "true" activity the size of 128 x 128 x 1. The encoder consists of 12 convolutional and three Dense layers encoding the sinogram to 2048 features.  The decoder is composed of 17 convolutional layers converting the output of the encoder to the final $128 \times 128$ activity distribution. 

The model was trained with 600,000 software phantoms, 90\% for training and 10\% for validation for 1000 epochs. Structural Similarity (SSIM) Index \cite{wang2004image} was used as shown in equation \ref{eq:SSMI} was used as the loss function

\begin{equation}
  SSIM(x,y) = \frac{(2\mu_x\mu_y + C_1)  (2 \sigma _{xy} + C_2)} 
    {(\mu_x^2 + \mu_y^2+C_1) (\sigma_x^2 + \sigma_y^2+C_2)}
  \label{eq:SSMI}
\end{equation}

where $\mu _{x}$ and $\mu _{y}$ are the average of  $x$ and $y$; $\sigma _{x}^{2}$ and $\sigma _{y}^{2}$ are the variance of $x$ and $y$; $\sigma _{{xy}}$ is the covariance of $x$ and $y$; $c_{1}=(k_{1}L)^{2}$, $c_{2}=(k_{2}L)^{2}$ two variables to stabilize the division with weak denominator; $L$ the dynamic range of the pixel-values (typically this is $2^{{\#bits\ per\ pixel}}-1)$; $k_{1}=0.01$ and $k_{2}=0.03$ by default.

\section{Results}
\label{sec:results}

To assess the performance of the proposed CNNR method versus existing methodologies, the Mean Square Error (MSE), Mean Absolute Error (MAE), Structural Similarity (SSIM) Index \cite{wang2004image}, and the Pearson Correlation Coefficient (PCC) \cite{benesty2009pearson} were used. As the results show, Table \ref{table:results}, the proposed CNNR method (highlighted in \textbf{bold}) outperforms all other methodologies, especially in the medium and high noise scenarios.

\begin{figure*}[htp]
    \centering
        \includegraphics[width=18cm, height=12cm]{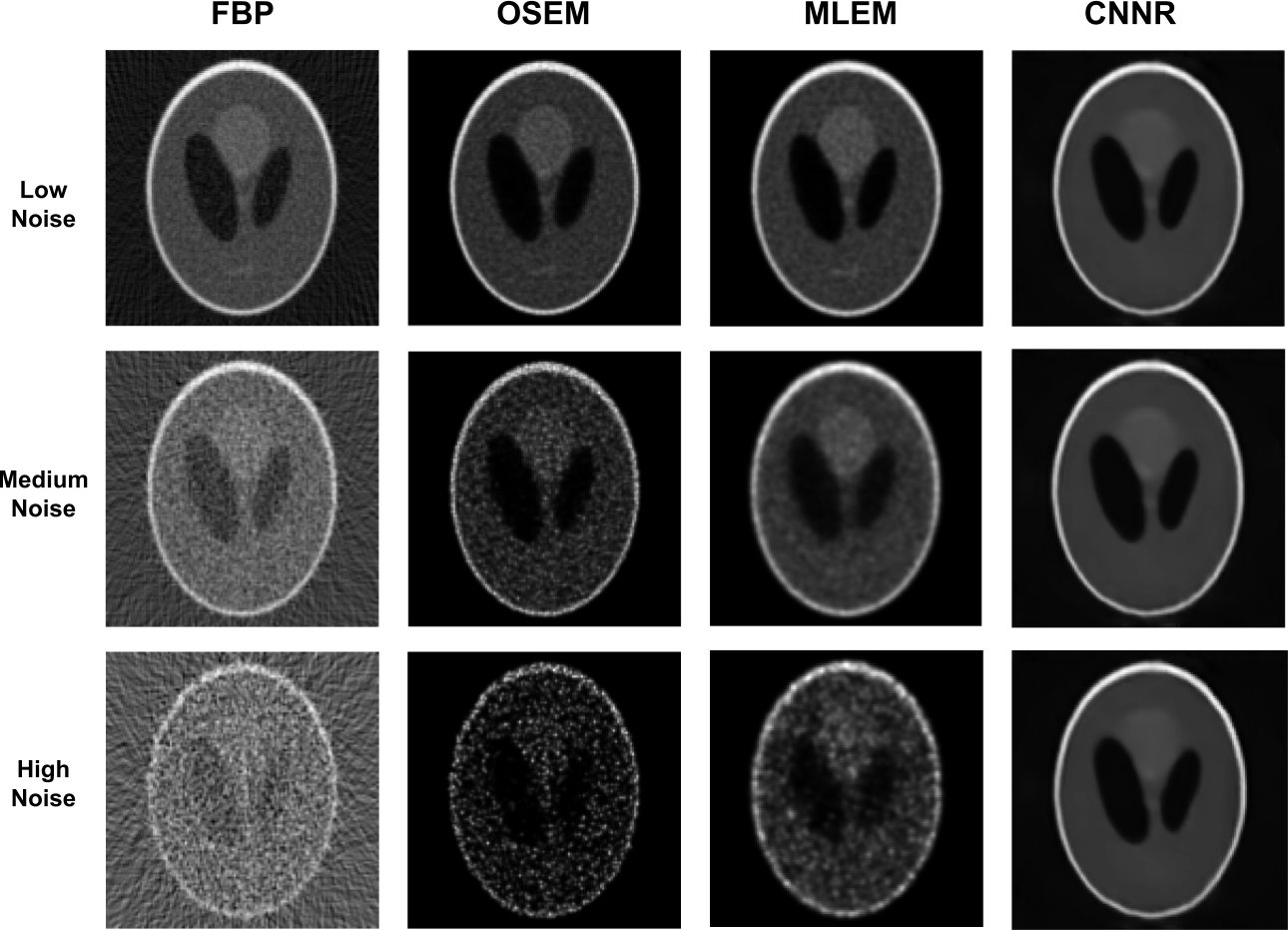}
    \caption{Evaluation and comparison of the proposed CNNR method versus the OSEM, MLEM and FBP methods. The results obtained using CNNR compare favourably to those obtained with the widely used FBP, OSEM and MLEM methods.}
    \label{fig:results}
\end{figure*}

\begin{table*}[ht]
\setlength\tabcolsep{8pt}

\renewcommand{\arraystretch}{1.5}
\centering
\caption{Results}
\label{table:results}
\begin{tabular}{|c | c | c | c | c | c | c | c | c | c | c | c | c |}
\hline
Methodology & \multicolumn{4}{c|}{Low Noise} & \multicolumn{4}{c|}{Medium Noise} & \multicolumn{4}{c|}{High Noise} \\
\hline
 & MSE & MAE & SSIM & PCC & MSE & MAE & SSIM & PCC & MSE & MAE & SSIM & PCC \\
 \hline
FBP & 0.003 & 0.034 & 0.834 & 0.963 & 0.009 & 0.068 & 0.514& 0.889 & 0.034 & 0.136 & 0.205& 0.561 \\
OSEM & 0.002 & 0.022 & 0.929 & 0.978 & 0.009 & 0.047 & 0.784 & 0.948 & 0.025 & 0.080 & 0.591 & 0.739 \\
MLEM & 0.003 & 0.022 & 0.928 & 0.980 & 0.006 & 0.034 & 0.849 & 0.957 & 0.010 & 0.049 & 0.736 & 0.862\\
\textbf{CNNR} & \textbf{0.002} & \textbf{0.021} & \textbf{0.938} & \textbf{0.981} & \textbf{0.003} & \textbf{0.023} & \textbf{0.938} & \textbf{0.971} & \textbf{0.004} & \textbf{0.026} & \textbf{0.930} & \textbf{0.962}\\

\hline
\end{tabular}
\end{table*}

\addtolength{\textheight}{-1. cm}
\section{Discussion and Conclusions}
\label{sec:conclussion}

In this paper, we propose a new method to perform SPECT  image reconstruction using convolutional neural networks and demonstrate the effectiveness of the method. For the training of the proposed convolutional neural network, we randomly created and used 600,000 software phantoms. 
Furthermore, we used existing methods, such as FBP, OSEM and MLEM applied to Shepp-Logan software phantoms \ref{fig:shepp_logan}, to compare with the proposed method to assess the performance of the method. The results of FBP, OSEM, MLEM and the proposed method are outlined Table \ref{table:results} and Figure \ref{fig:results}. The proposed CNNR method outperforms all other methods in reconstructing the Shepp-Logan software phantoms, as the results show, particularly in medium and high noise conditions. Even though the conditions the proposed method is tested and the results presented are suitable to demonstrate the proposed CNNR method capabilities, compared to existing methods, additional experimentation is needed with real phantoms. These experimentations will evaluate the implied use of the proposed method in clinical studies.

\bibliography{main2} 
\bibliographystyle{ieeetr}

\end{document}